\documentclass[letterpaper, 10 pt, conference]{ieeeconf}  

\IEEEoverridecommandlockouts       
\usepackage{style}

\title{\LARGE \bf
Learning Safe Humanoid Navigation from Reduced Order Models
}

\author{William D. Compton$^1$, Zachary Olkin$^1$, Ryan Bena$^2$, Aaron D. Ames$^{1,2}$ 
\thanks{$^1$The authors are with the Department of Computing and Mathematical Sciences, California Institute of Technology, Pasadena, CA.}
\thanks{$^2$The authors are with the Amazon Safe Autonomy Frontiers (SAF) Lab.}
\thanks{This research is supported by Technology Innovation Institute (TII).}
}

\begin{document}
\bstctlcite{IEEEexample:BSTcontrol}

\maketitle
\thispagestyle{empty}
\pagestyle{empty}

\begin{abstract}
Research in humanoid robotics has achieved rapid progress in locomotion, and recent results have pushed the boundary on autonomous navigation. 
We demonstrate that a standard single-stage RL navigation pipeline struggles to scale to multi-level and multi-story terrain, limited by the difficulty of complex humanoid terrain interactions such as stairs.
To overcome this challenge, we decompose the navigation problem into two pieces. 
First, we train a policy operating on the reduced order dynamics but with full 3D LiDAR observations to navigate complex, multi-story terrain.
We then utilize this navigation knowledge to kickstart a policy operating on the full-order humanoid dynamics, with a frozen locomotion policy in the loop.
Additionally, we demonstrate that applying a Poisson safety filter to the navigation policy output recovers safety in the presence of out-of-distribution obstacles, without dropping navigation success rate. 
We demonstrate the resulting \mbox{RoM-Nav} policy on a Unitree G1, accomplishing mapless multi-floor navigation covering trials with over 10\,m of vertical displacement and over 100\,m of path length.
Project page with videos \url{https://wdc3iii.github.io/rom-nav/}.
\end{abstract}

\section{Introduction}
\begin{figure}[t]
\centering
\includegraphics[]{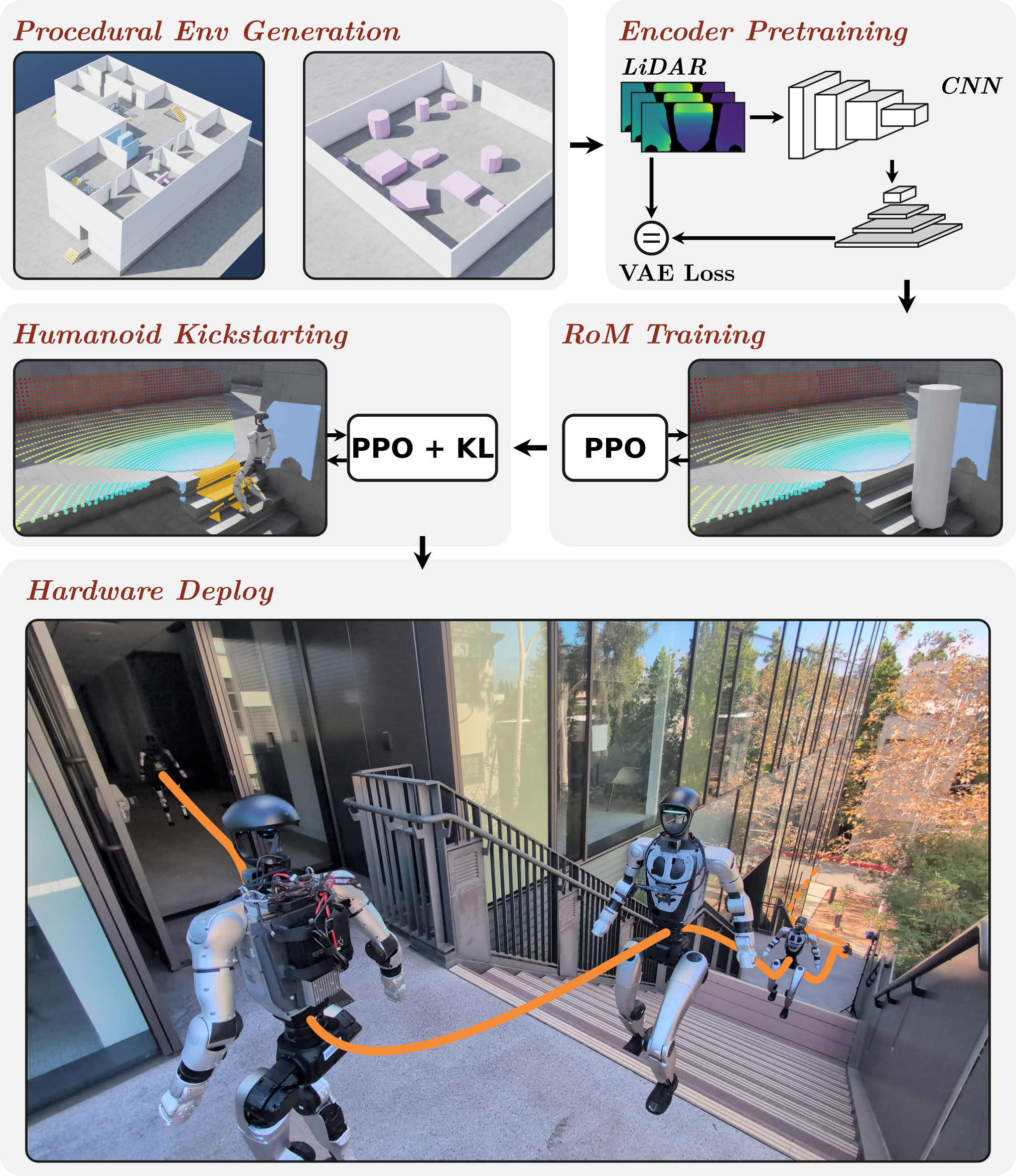}
\caption{Training and deployment of the RoM-Nav policy for mapless multi-floor humanoid navigation. The humanoid autonomously traverses stairs covering multiple floors (over 10\,m of vertical displacement), and long-horizon, highly non-convex paths, exceeding 100\,m in length.}
\label{fig:hero}
\vspace{-5mm}
\end{figure}

Safe navigation of complex, real-world environments is a fundamental capability for humanoid robots. 
Any task requiring a robot to have legs almost certainly contains a navigation component, and will likely challenge the robot to handle terrain that is not flat, such as sidewalk curbs or staircases.
Recent advances in perceptive humanoid locomotion have unlocked impressive new capabilities, including running \cite{olkin2026chasing}, stairs \cite{zhang2026rpl, compton2026terrain}, stepping stones \cite{wang2025beamdojo}, parkour \cite{wu2026perceptive}, and loco-manipulation \cite{liu2025opt2skill}.
As locomotion controllers become more agile, navigation planning becomes increasingly nontrivial; success becomes intertwined with the capabilities of the locomotion controller, which can be difficult to represent in a useful way for the navigation planner. 

Reinforcement learning (RL) based navigation has provided a way to meet this challenge.
RL provides many benefits; its lightweight runtime computation removes the complexity of real-time optimization.
It can integrate directly with sensor data, removing the need for occupancy or traversability estimation. 
It learns directly to handle complex and terrain-dependent system dynamics, rather than requiring an explicit model or capability envelope.
However, it comes with its own set of challenges. Sparse rewards and long-horizon tasks can lead to training instability or exploration collapse. 
Limitations in useful memory can handicap long-horizon performance, and sim-to-real gaps can lead to safety violation, especially if training geometry is simplistic.

To address these limitations, we propose a two-stage training paradigm, where a navigation policy is first trained on a reduced order model (RoM) of the humanoid.
By trivializing the complexity of the environment interaction, the policy trained on the reduced order model quickly achieves superior performance to single-stage navigation policies. 
Then, with a combined objective of matching the RoM policy's action distribution and a standard PPO objective, we use the navigation capabilities of the RoM to kickstart a policy which handles the complexities of the terrain interaction and full-order dynamics of the humanoid under its frozen locomotion policy. 
We demonstrate improvement over single-stage training and show the gap surfaces almost entirely in problems requiring significant z-height conditioning, for instance cross-floor goals in multi-story buildings. 

Finally, we demonstrate these navigation behaviors on hardware to navigate complex real-world terrain, both outdoors and in multi-floor buildings.
Critically, we leverage Poisson safety filters \cite{bena2025geometry, yamaguchi2026layered} online to bridge environmental sim-to-real gaps, such as complex obstacle geometries, which are difficult to capture during training for an unknown deployment environment.  
We show that Poisson safety filters ensure safety when encountering out-of-distribution (OOD) obstacle geometry, while maintaining navigation success rate. 

\begin{figure*}[t]
\centering
\includegraphics[]{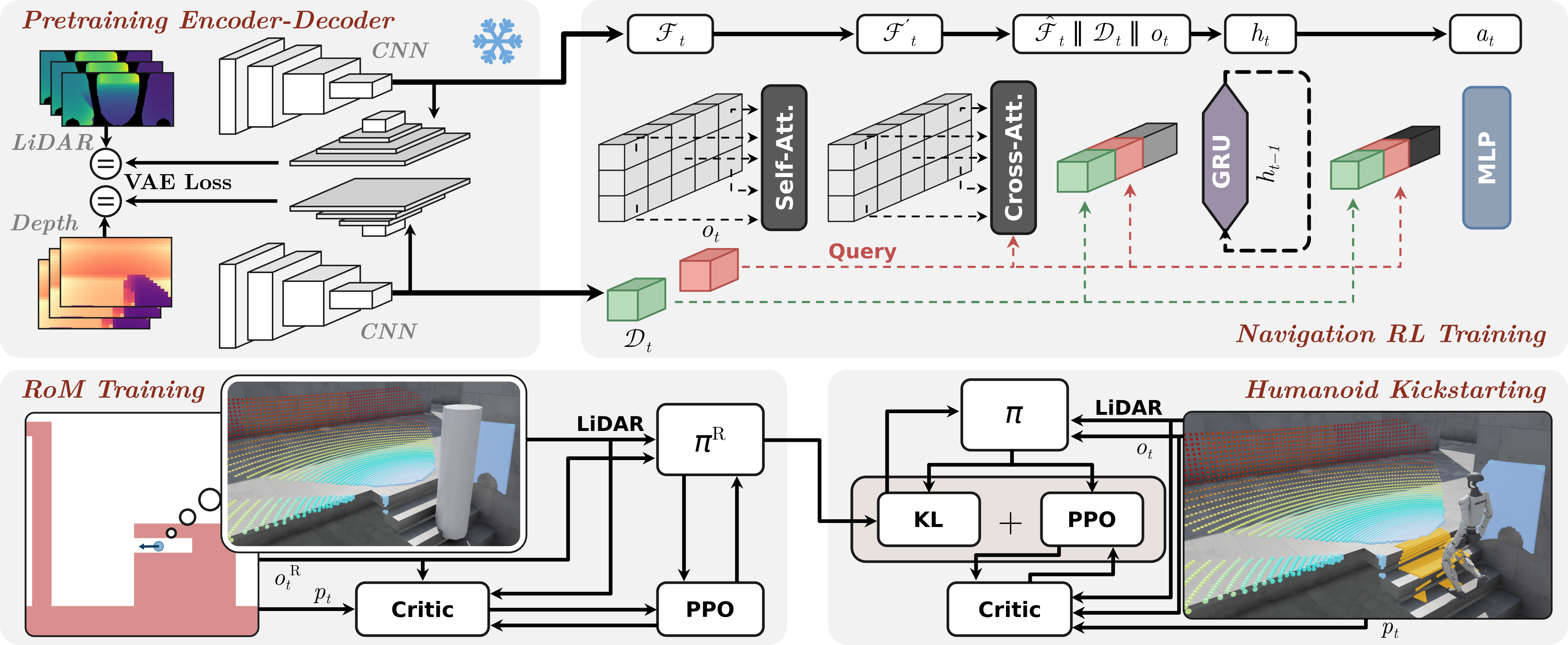}
\caption{RoM-Nav architecture. LiDAR and Depth inputs are processed via a pretrained CNN encoder. LiDAR passes through self- then cross-attention. The resulting embeddings are processed by a GRU unit and mapped to actions via an MLP head. A RoM policy is trained first, which is then used to kickstart the RoM-Nav policy by balancing KL divergence with a PPO objective on the humanoid with frozen locomotion controller.}
\label{fig:arch}
\vspace{-5mm}
\end{figure*}

Concretely, our contributions are as follows:
\begin{itemize}
    \item \textbf{Multi-Story Mapless Humanoid Navigation.} We deploy a mapless RL navigation policy on hardware to safely accomplish challenging navigation problems in real-world environments, including outdoor deployment of 100\,m, and indoor tasks spanning multiple floors. To our knowledge, this is the first published humanoid navigation work to demonstrate learned mapless navigation for cross-floor goals in multi-story buildings. 
    \item \textbf{Reduced Order Model Kickstarting for Navigation.} We propose a two-stage navigation reinforcement learning pipeline (\cref{fig:arch}): a policy is first trained to navigate a RoM, which kickstarts a policy operating on the humanoid. We demonstrate effective spawn/goal sampling and LiDAR encoder pretraining which unlock multi-story navigation capabilities.
    \item \textbf{Online Poisson Safety Filter for OOD Obstacles.} We demonstrate degradation of the policy when faced with out-of-distribution obstacles, and leverage the Poisson safety filter to cheaply ensure safety on hardware while maintaining navigation success rate.
\end{itemize}

\section{Related Work}

We review navigation for humanoids, learned navigation for legged robots, the choice of dynamics to train against, and runtime safety for learned policies.
 
\textbf{Humanoid Navigation:} Early work in humanoid navigation relied on model-based methods, be it kinematic \cite{griffin2019footstep}, dynamic \cite{dai2014whole}, reduced order \cite{narkhede2022sequential}, or sampling-based \cite{huang2023efficient}.
These methods typically perform heavy preprocessing of the environment, such as floor and obstacle height maps \cite{gutmann20083d} or convex steppable regions to optimize over \cite{deits2014footstep}.
Modern extensions of these ideas have injected learned elements such as traversability estimates \cite{lin2021long, frey2023fast, yoon2025state} or even forward dynamics models \cite{roth2025learned} to relieve the perception bottleneck, while maintaining the model-based backbone.
These methods address local navigation, limited by planning horizon and the convexity of the regions they optimize over, unless paired with waypoints or a global map \cite{huang2023efficient, lin2021long}, as in recent work navigating stairs on a humanoid \cite{compton2026terrain}.
Rather than optimize over a preprocessed environment, a large body of work learns navigation commands directly from sensor data.
For humanoids, these efforts have targeted local navigation, such as goal-conditioned locomotion across constrained height-varying terrain \cite{ben2026gallant}, waypoint-guided obstacle avoidance on stairs and slopes \cite{zhang2026focusnav}, or teacher-distilled obstacle avoidance \cite{han2026guidewalk}.
Vision-language models have also reached humanoids, but interface through parameterized motion primitives \cite{cheng2024navila} or waypoints for a local planner over a SLAM map \cite{du2025vl}, and so inherit the same local or map-based structure.
Humanoid navigation to date, model-based or learned, has therefore been primarily local or reliant on a map.
 
\textbf{Learned Navigation:} Learned navigation is considerably more mature on quadruped and wheeled-legged robots, with significant work spanning from training navigation and locomotion simultaneously and end-to-end \cite{rudin2022advanced}, to hierarchical policies commanding a separately trained locomotion controller \cite{hoeller2021learning, truong2021learning, fu2022coupling}, to long-horizon deployments with explicit memory \cite{lee2024learning} or specialized recurrent structures \cite{yang2025spatially}.
Our single-stage pipeline follows this last line of work.
Across all of these learned approaches, humanoid and quadruped alike, the training environments contain a single traversable level, so that the height of the goal carries no information its planar position does not; where stairs and slopes appear, they are terrain features to be crossed en route.
Spawns and goals are sampled uniformly over the traversable area, and curricula, where present, scale terrain difficulty rather than goal placement \cite{rudin2022advanced}.
To our knowledge, no learned, mapless humanoid navigation policy has been commanded to a goal on a different floor of a building.
 
\textbf{Navigation Dynamics Model:} These works also differ in the fidelity of the dynamics the navigation policy is trained against.
Navigation at scale was first achieved on a point robot with discrete actions \cite{wijmans2019dd}, and imperative planners learn paths over a cost map without simulating the robot's dynamics \cite{yang2023iplanner, roth2024viplanner}.
Truong et al.\ make the case explicitly, showing that lower fidelity simulation yields higher sim-to-real transfer for navigation on a quadruped \cite{truong2023rethinking}, and kinematically trained navigation policies have since been paired zero-shot with learned locomotion controllers \cite{kareer2023vinl}.
While these works transfer zero-shot, they are evaluated in environments without complex terrain or multiple stories; we find that the low-fidelity approach works for same-level goals but breaks for cross-floor goals that require non-trivial terrain traversal (\cref{sec:results}), and so add a second training stage on the full humanoid dynamics.
NavRL++ also trains in multiple stages, but its secondary stages introduce sensing and actuation perturbations for sim-to-real transfer rather than higher-fidelity dynamics \cite{xu2026navrl}.
Teacher-student training in legged robotics is typically observation-privileged, distilling a teacher with privileged terrain and state information into a student with onboard sensing \cite{miki2022learning}; we instead kickstart \cite{schmitt2018kickstarting} from a dynamics-privileged teacher, whose observations are a subset of the student's but which was trained on a simpler system.
 
\textbf{Safety Filters:} Regardless of the training regime, learned navigation policies carry no collision guarantee, and runtime safety filters are commonly layered on top of them.
For legged navigation, learned reach-avoid value functions have filtered high-speed quadruped policies \cite{he2024agile}, reachability-based filters with disturbance estimation have been deployed from LiDAR in unknown environments \cite{lin2025one}, and velocity-obstacle shields have filtered RL policy outputs on aerial and legged platforms given tracked obstacles \cite{xu2026navrl}.
Each depends on a learned value function, a disturbance model, or an obstacle detector.
Poisson safety functions synthesize a control barrier function directly from raw occupancy, requiring none of these, and have been paired with model-based planners on humanoids \cite{bena2025geometry, yamaguchi2026layered}.
We evaluate this filter on a learned humanoid navigation policy against obstacles chosen to lie outside its training distribution.

\section{Methods}

The navigation policy is trained for deployment on a Unitree G1 equipped with a Mid-360 LiDAR and a downward-facing ZED Mini depth camera. The frozen locomotion policy for the robot is trained using LIP-CLF RL \cite{compton2026terrain}. 

\subsection{Pretraining Vision Encoders}

The LiDAR range image and depth image are both high-dimensional inputs which require significant visual processing to extract useful information.
Rather than extract these features during the RL training, we pretrain a CNN encoder for both the range image and depth image using a standard VAE approach \cite{kingma2013auto}.
Encoder pretraining for depth has been used in navigation RL \cite{hoeller2021learning, yang2025spatially}; the LiDAR case differs in FoV and noise structure and has not been ablated in a humanoid navigation context.
The VAE contains a set of convolution heads to project the image down to a Gaussian latent space.
A set of deconvolution layers maps back to a full resolution image.
We train heads to accomplish depth, XYZ, and valid mask reconstruction, and ground/obstacle/stair/ramp segmentation, to ensure retention of critical features in the VAE latent.
Depth and XYZ reconstruction use an MSE loss; masking and segmentation use BCE.
The VAEs are trained with synthetic images sampled from randomized robot poses. Stairs and ramps are over-sampled due to their spatial sparsity in the tiles. 
During VAE training, the inputs are noised, pixel dropout between 0--80\% is applied, and variations on the robot's head occlusion mask are applied.
VAEs are trained for 10 epochs on 1M uniform and 1M stair, ramp oversampled images. After training the encoder is frozen in navigation policy, as depicted in \cref{fig:arch}.

\subsection{Policy Architecture}

The observations passed to the navigation policy are a LiDAR range image, a depth image, and a nonvisual observation.
The range image bins the LiDAR into $1^\circ$ angular pixels containing distance, $x,y,z$ in the LiDAR frame, and validity channels. 
The depth image is down-sampled from the ZED Mini depth to $26 \times 30$.
The goal observation is encoded as $o_g = [d_x, d_y, \log \|d\| / \log d_{\text{max}}, d_z, \sin(\gamma), \cos(\gamma)]$ where $d \in \R^3$ is the relative location of the goal and $\gamma$ is the relative heading between the goal heading and the robot heading. 
The non-visual observation is $o_t = [o_g, \tau, \omega, p_g, a_{t-1}]$, with $\tau$ the locomotion phase, $\omega$ and $p_g$ the angular rate and projected gravity of the pelvis link, and $a_{t-1}$, the previous action.

The policy architecture, pictured in \cref{fig:arch}, leverages two vision encoders for the LiDAR range image and depth image.
The LiDAR passes through a frozen CNN encoder, a layer of self-attention, and a layer of cross-attention with four learned queries and one query encoding the nonvisual observations. 
The depth passes through a pretrained CNN encoder and is flattened. 
The vision embeddings are concatenated with the nonvisual encoding, and fed into a two-layer GRU, whose recurrent structure gives the policy memory to compensate for partial observability and to support the exploration required for navigation problems. 
The resulting recurrent embedding $h_t$ is concatenated again with the visual and nonvisual embeddings, to pass through a dense layer mapping to velocity outputs, parameterized as Beta distributions, a common bounded distribution for navigation RL \cite{lee2024learning}.

We claim no contribution in architecture; ours follows \cite{yang2025spatially}.
We use a GRU in place of the SRU, as the GRU is more common and preliminary experiments found no practical difference, likely due to the wider FoV of the LiDAR.

\subsection{RoM Kickstarting}
The proposed method kickstarts \cite{schmitt2018kickstarting} the final navigation policy from one trained on a single integrator with heading, depicted in the bottom half of \cref{fig:arch}, with dynamics:
\begin{equation*}
    \begin{bmatrix}
        x_{t+1} \\ y_{t+1} \\ \theta_{t+1}
    \end{bmatrix} = \begin{bmatrix}
        x_t + (v_{x,t} \cos{\theta_t} - v_{y,t} \sin{\theta_t})\Delta t \\
        y_t + (v_{y,t} \cos{\theta_t} + v_{x,t} \sin{\theta_t})\Delta t \\
        \theta_t + \omega_{z,t} \Delta t
    \end{bmatrix}.
\end{equation*}
The environment for the single integrator is an occupancy grid derived from the 3D environment for the humanoid (see \Cref{sec:envs}).
When a command leads to an occupied cell, the penetrating component is projected out, sliding along the occupied boundary.
The RoM's LiDAR is placed at the humanoid's standing pose above the terrain.

The RoM trains with only LiDAR, goal, and previous action, the signals relevant to RoM navigation. 
Including the depth observation made no difference in performance, likely as the RoM dynamics are not terrain-dependent. 
The RoM policy is trained for 2000 iterations using PPO, with 4096 environments on one H100 GPU in 12 hours.

Once the RoM policy is trained, it kickstarts a policy using the architecture in \cref{fig:arch}.
Kickstarting uses a weighted combination of the PPO objective and the KL divergence between the RoM policy and the RoM-Nav policy \cite{schmitt2018kickstarting},
\begin{equation*}
    \mathcal{L} = \mathcal{L}_{\mathrm{PPO}} + \lambda \mathcal{L}_{\mathrm{KL}}(\pi, \pi^R).
\end{equation*}
The weight $\lambda$ is $1$ for the first 100 iterations, then decays to $0.05$ by 1100 iterations, where it holds until training is finished at 2000 iterations.
The RoM-Nav policy is trained with 4096 environments on a single H100 GPU, taking 32 hours. The combined training is under 45 hours on one GPU.

\begin{table}[t]
\centering
\caption{Navigation reward terms. $^\dagger$ indicates geodesic gating.}
\label{tab:nav-rewards}
\footnotesize
\setlength{\tabcolsep}{3pt}
\begin{tabular}{@{}lrl@{}}
\toprule
Term & $w$ & Per-step value $r$ \\
\midrule
\multicolumn{3}{@{}l}{\emph{Goal tracking} (dense)} \\
position$^\dagger$       & $0.2$ & $1 + \exp(-d_{xy}^2 / 0.5^2) + \exp(-d_{xy}^2 / 0.1^2)$ \\
heading$^\dagger$ & $0.2$ & $\exp(-e_\psi^2 / 0.5^2)$ \\
height                                & $0.2$ & $\exp(-z_e^2 / 2.5^2) + \exp(-z_e^2 / 0.5^2)$ \\
stand$^\dagger$   & $0.2$ & $\exp(-\lVert a_t \rVert^2 / 0.1^2)$ \\
\addlinespace
\multicolumn{3}{@{}l}{\emph{Progress shaping} (CLF, best-gated)} \\
geodesic & $0.1$ & $\big[\tfrac{\Delta d_{\mathrm{geo}}}{\min\{0.8d_{\mathrm{geo}},\, \overline{v}_{xy}\Delta t\}}\big]_0^1$ \\
height   & $0.1$ & $\big[\tfrac{\Delta z_e}{\overline{v}_z \Delta t}\big]_0^1$ \\
\addlinespace
\multicolumn{3}{@{}l}{\emph{Penalties}} \\
collision    & $-1$  & \# bodies in contact \\
action rate  & $-0.01 \to -0.1$ & $\lVert a_t - a_{t-1}\rVert^2$ \\
\bottomrule
\end{tabular}
\vspace{-5mm}
\end{table}

\subsection{RL Environment Configuration} \label{sec:envs}
In this section, we provide the details of the training environments. 
The navigation policies run at 5\,Hz. The frozen locomotion policy runs at 50\,Hz, with velocity limits of 1\,m/s forward, 0.25\,m/s lateral, and 1\,rad/s angular.
The LiDAR FoV is set to a forward-centered $220^\circ$, to facilitate human support during testing which occludes the LiDAR. Ultimately, all hardware experiments are run with no support.

\textbf{Rewards:}
The rewards follow the structure of established literature \cite{yang2025spatially, xu2026navrl, wang2026guide}, consisting of goal tracking rewards, progress shaping rewards, and penalties, shown in \cref{tab:nav-rewards}, where $d_{xy}$ is the planar distance to goal, $e_\psi$ is the heading error, $z_e$ the vertical error, $\overline{v}_{xy}$ is the speed required for maximum geodesic progress reward, and $\overline{v}_z$ is the vertical speed required for maximum height progress reward. 
The two primary additions to a standard navigation reward set are the height goal tracking reward and height progress reward, since the policy is trained on multi-floor buildings.
The gated rewards only fire when within 2\,m of geodesic distance to the goal.
The progress shaping rewards are input-limited CLF rewards \cite{li2026clf}, which incentivize progress to the goal along the geodesic (in $x,y$) and along height. These rewards only fire when at the episode best, to avoid circling behaviors accumulating free progress. While the CLF reward provides a modest training stabilization influence, the height rewards are critical to the success in multi-floor environments. The action rate penalty changes at iteration 1000, allowing exploration early but encouraging smoothness late.

\begin{figure}[t]
\centering
\includegraphics[]{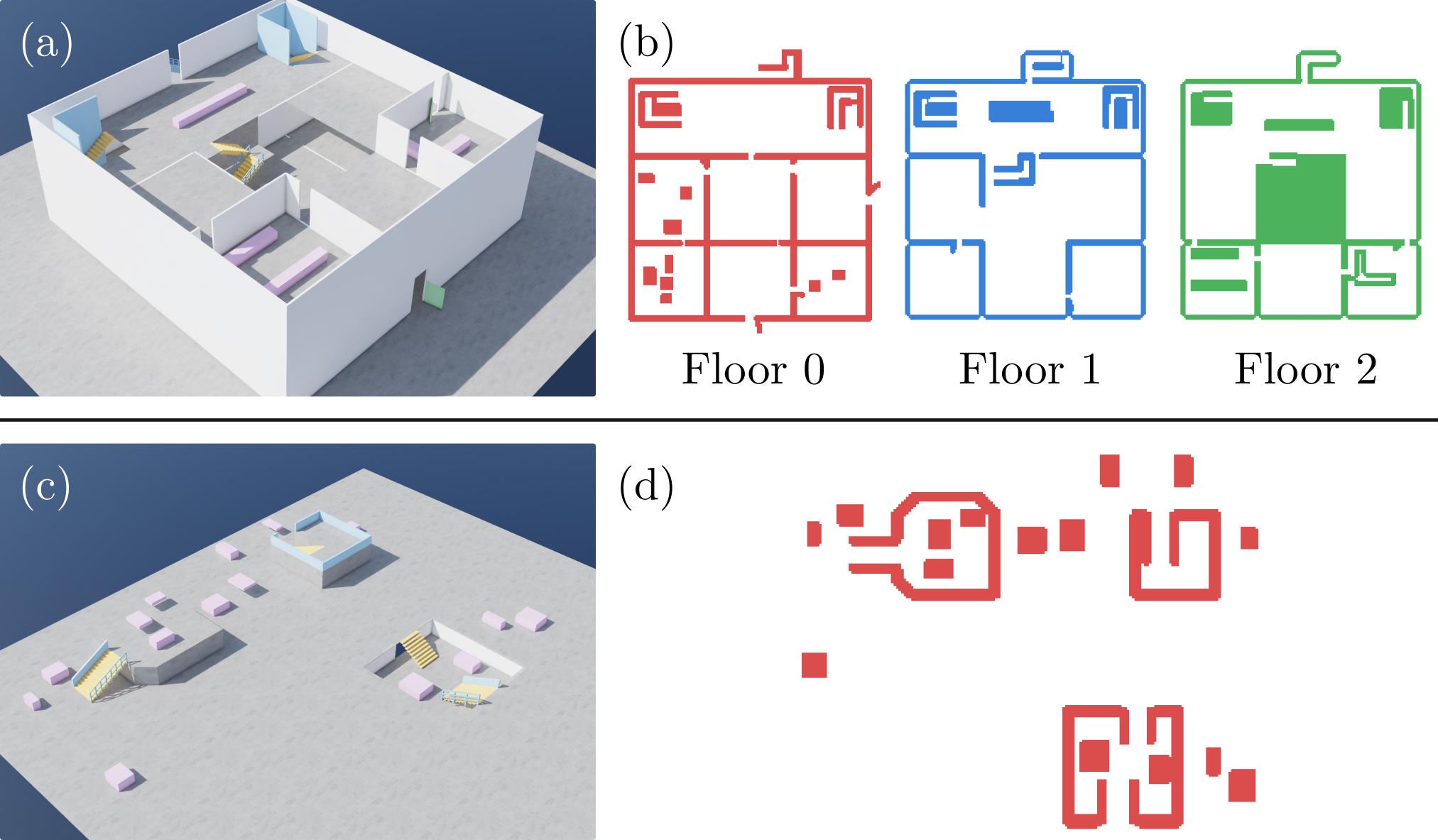}
\caption{Environment tiles. (a) multi-story building and (b) its occupancy grid. (c) multi-level outdoor environment and (d) its occupancy grid.}
\label{fig:envs}
\vspace{-5mm}
\end{figure}

\textbf{Environments:} 
Where existing works utilize single-level layouts, our deployment in complex multi-level real-world environments requires a different approach. 
We enumerate several classes of procedurally generated tiles, including outdoor tiles, single room tiles, and multi-story building tiles. 
A tile is considered to be ``multi-story'' if there exist $x,y$ coordinates which contain multiple disjoint $z$ locations which are coherent goals, i.e., the same $x,y$ position on different floors.
A few sample tiles are shown in the left panes of \cref{fig:envs}.
As the tiles are generated programmatically, and since the locomotion policy was trained to locomote over specific terrain, ground truth traversability is easily recorded during terrain generation. 
Obstacle and traversability information is rasterized into a 0.2\,m occupancy grid during terrain generation, seen in the right panes of \cref{fig:envs}. 
This occupancy map is then used as the collision model for the RoM during its training. The occupancy map is maintained per floor for the multi-floor environments, where the robot transitions between floors based on a height threshold. The occupancy grids are locally consistent in floor transition regions. 

\textbf{Spawn/Goal Sampling:}
Most existing works sample spawns and goals uniformly in their environments, perhaps under a difficulty curriculum on the environment. 
As discussed in \Cref{sec:results_rlchoices}, we found naive sampling of the spawns and goals made z-height dependencies in multi-story environments difficult to learn. 
To address this, we adopt a specialized reset distribution over uniformly distributed, guaranteed free space spawn and goal locations.
First, with probability $p=0.3$ the spawn is selected to be on or near the mouth of a stair or ramp. To place initial conditions on the stair or ramp, where a nominal standing initial condition does not work well, a gait library is collected of the locomotion policy traversing stairs or ramps under standard commands. The closest stair or ramp in the gait library is selected, and the robot is placed at a random time along the gait from the library. When this specialized reset is not sampled, the fallback is uniform sampling over the valid spawns. 

The goal sampling is geodesically capped and cross-level biased. 
A wavefront algorithm starting at the goal computes free-space geodesic distance to each initial condition. 
Goals beyond 30\,m geodesic distance are masked out, ensuring feasibility during the 45\,s episode with 1\,m/s max speed. 
This goal mask computation is parallelized via a CUDA kernel, taking 14 minutes to precompute for the entire environment. 
Second, goals are biased to cross height levels. 
Three story buildings tiles sample cross-level goals 100\% of the time; other tiles, 50\%. 
This bias helps prevent under-conditioning on $z$, as cross-level goals require traversal of difficult terrain and longer horizons. 

\begin{table}[t]
\centering
\caption{Navigation performance across training methods. SR@45 / SR@120 are success within 45\,s (the training budget) and 120\,s; T2G is mean time to reach the goal over successful episodes; SPL is success weighted by path length against the geodesic route. \textbf{Bold} marks the best humanoid arm per column. $^*$ RoM (cyl) is evaluated on RoM dynamics.}
\label{tab:nav-perf}
\small
\setlength{\tabcolsep}{5pt}
\begin{tabular}{@{}lrrrr@{}}
\toprule
Arm & SR@45 (\%) & SR@120 (\%) & T2G (s) & SPL \\
\midrule
RoM (cyl)$^*$      & 84.1 & 93.6 & 25.6 & 0.776 \\
\addlinespace
Single-Stage           & 62.2 & 81.7 & 32.1 & 0.690 \\
RoM (hum)      & 74.2 & 88.1 & 29.8 & 0.713 \\
RoM-Nav      & \textbf{82.3} & \textbf{92.8} & \textbf{28.8} & \textbf{0.769} \\
\bottomrule
\end{tabular}
\vspace{-5mm}
\end{table}

\subsection{Poisson Safety Filter}

The Poisson safety filter modifies velocity commands coming from the navigation policy before passing them to the locomotion policy, to avoid colliding with OOD obstacles.
We adopt the Real-Time CBF-QP layer deployed in \cite{yamaguchi2026layered}, which we summarize here.
The Poisson safety filter uses a single integrator approximation, with state $p = [x, y]$ and input $v=[v_x, v_y]$ satisfying dynamics $\dot{x} = v_x$, $\dot{y} = v_y$. 

According to Control Barrier Function (CBF) theory \cite{ames2016control}, a smooth function $h(p)$, positive in free space and negative in occupied space, is a control barrier function if, for $\alpha > 0$: 
\begin{equation*}
    \sup_{v} \dot{h}(p, v) \geq -\alpha h(p).
\end{equation*}
Furthermore, selecting $v$ such that $\dot{h}(p, v) \geq -\alpha h(p)$ ensures that $h(p) \geq 0$ for all time. 
To construct $h$, the point cloud is rasterized into a 2D occupancy grid (0.05\,m), after removing the ground. 
The occupancy is convolved with a circle of the robot's radius, and the boundary of occupied space is identified.
Then Poisson's equation is solved over this grid, enforcing $h(p) = 0$ on the boundary and positive divergence in free space via a forcing function; see \cite{bena2025geometry, yamaguchi2026layered} for details.

Once the Poisson safety function $h$ has been synthesized, the Real-Time CBF-QP finds the minimum deviation to the desired input $v_{\mathrm{des}}$, output by the RoM-Nav policy, which satisfies the barrier constraint, with $\alpha=0.75$:
\begin{equation*}
    v_{\mathrm{safe}} = \arg \min_{v} \|v - v_{\mathrm{des}} \|^2 \quad \mathrm{s.t.} \quad \frac{d h}{d p} \big\vert_p v \geq -\alpha h(p).
\end{equation*}
This safe planar velocity is then passed to the locomotion policy, avoiding collision in the presence of OOD obstacles. 

\begin{table}[t]
\centering
\caption{Success rate paired contrasts by same-level vs. cross-level goals. Positive favors RoM-Nav; $\dagger$ marks a 95\% confidence interval excluding zero.}
\label{tab:nav-levels}
\footnotesize
\setlength{\tabcolsep}{4pt}
\begin{tabular}{@{}lcc@{\hspace{1.1em}}cc@{}}
\toprule
& \multicolumn{2}{c}{45\,s} & \multicolumn{2}{c}{120\,s} \\
\cmidrule(lr){2-3}\cmidrule(lr){4-5}
Contrast (\%) & same & cross & same & cross \\
\midrule
RoM-Nav\ $-$ RoM (hum) & $+1.5$ & $+15.2$\rlap{\textsuperscript{$\dagger$}} & $+0.6$ & $+9.1$\rlap{\textsuperscript{$\dagger$}} \\
RoM-Nav\ $-$ Single-Stage & $+6.8$\rlap{\textsuperscript{$\dagger$}} & $+34.6$\rlap{\textsuperscript{$\dagger$}} & $+3.0$\rlap{\textsuperscript{$\dagger$}} & $+19.7$\rlap{\textsuperscript{$\dagger$}} \\
\bottomrule
\end{tabular}
\vspace{-5mm}
\end{table}

\section{Results} \label{sec:results}

In this section we examine the RoM-Nav method, showing its improvement over the single-stage training approach, and justify the pretrained encoders and spawn/goal sampling.
We demonstrate the effectiveness of the Poisson safety filter for safety around OOD obstacles on hardware.
Finally, we deploy the method in real-world multi-floor environments. 

\subsection{RoM Kickstarting}
First, we evaluate the RoM-Nav method. 
We compare a few methods: Single-Stage, trained directly on the humanoid via PPO for 4000 iterations. RoM (cyl) is the RoM policy evaluated \textit{on the RoM}, and provides an upper bound for the RoM-Nav policy. RoM (hum) is the RoM policy evaluated \textit{on the humanoid}. RoM-Nav is the proposed method, where the RoM policy kickstarts a policy trained on the humanoid, using the combined KL and PPO loss. 

\cref{tab:nav-perf} compares the performance of these policies on a paired test. Each policy is spawned from the same 1024 initial conditions, in a randomized terrain generated with a different seed from training. 
The metrics recorded are success rate, where success is reaching within 0.5\,m of the goal, in the training budget of 45\,s (SR@45), or within 120\,s (SR@120).
The longer evaluation allows the policy to take and recover from more wrong turns, reasonable in a mapless context. 
We also report mean time to goal (T2G) over successful trials, and 120\,s success weighted by inverse path length (SPL), a metric introduced in \cite{anderson2018evaluation} for evaluating navigation efficiency:
\begin{equation*}
\mathrm{SPL} = \frac{1}{N} \sum_{i=1}^{N} S_i
    \frac{\ell_i}{\max(p_i,\, \ell_i)},
\end{equation*}
where $S_i$ is the success of the $i$th trial, $p_i$ is the length of the traversed path, and $\ell_i$ the geodesic length.

Evaluating \cref{tab:nav-perf}, we see Single-Stage under-performing, while RoM-Nav performs within 1--2\% of the RoM (cyl). The RoM deploys reasonably well onto the humanoid without the kickstarting step, but kickstarting recovers an additional 8\% success at 45\,s, and an additional 5\% at 120\,s. \cref{tab:nav-levels} breaks these success results down by spawn/goal level, relative to the RoM-Nav policy. RoM-Nav makes all of its progress back in the cross-level trials, where stairs or ramps must be traversed to change $z$ level. Furthermore, the cross-level improvement primarily comes from fewer falls. On the 120\,s cross-level trials, the timeout rates are 4.9\% for RoM (hum) and 3.7\% for RoM-Nav, but the fall rate drops from 16.3\% on RoM (hum) to 8.3\% for RoM-Nav. The kickstarting leverages the RoM navigation ability, and robustifies it against the complex terrain interactions necessary to cross floors.

\subsection{RL Method Choices} \label{sec:results_rlchoices}
\begin{figure}[t]
\centering
\includegraphics[]{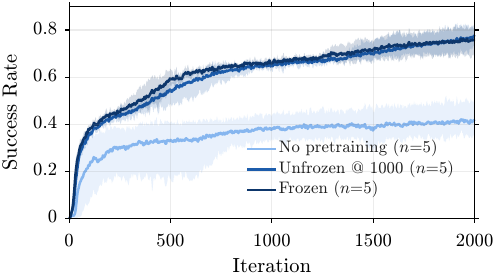}
\caption{An evaluation of the LiDAR encoder pretraining. Success rate vs. training iteration is plotted for three pretraining variations; mean over five random seeds, min-max shaded.}
\label{fig:pretrain}
\vspace{-2mm}
\end{figure}

\textbf{Pretrained Encoders:} 
To demonstrate the impact of the encoder pretraining, we train three variants of the RoM policy, using different LiDAR encoder pretrainings. No Pretraining randomly initializes the CNN and trains it with PPO. Unfrozen @ 1000 takes the pretrained CNN weights and unfreezes them only after the first 1000 iterations of PPO. To train the CNN weights on the H100 GPU, we reduce the environment count to 3072 and we double the PPO minibatches from 8 to 16. Frozen is a control arm, with the CNN weights frozen for all of training (same as deployed variant), but with the reduced environments and increased minibatches. 
\cref{fig:pretrain} shows the success rate training curves. Without pretraining, success is dramatically reduced. Frozen and Unfrozen @ 1000 reach equivalent performance. Additionally, we note that training with the unfrozen CNN weights increases the iteration time by 15\%. Furthermore, the RoM policy from \cref{tab:nav-perf} has a success rate 4\% higher than the best curve on this plot, due to its higher environment count and larger PPO batch. These results justify the frozen pretrained encoder.

\begin{table}[t]
\centering
\caption{Spawn/goal-distribution ablation on the RoM. Metrics identical to \cref{tab:nav-perf}. \textbf{Bold} marks the best arm per column (T2G excluded, as this metric skews when only short episodes are completed).}
\label{tab:spawngoal-perf-rom}
\small
\setlength{\tabcolsep}{4pt}
\begin{tabular}{@{}lrrrr@{}}
\toprule
Arm & SR@45 (\%) & SR@120 (\%) & T2G (s) & SPL \\
\midrule
RoM & \textbf{84.1} & \textbf{93.6} & 25.6 & \textbf{0.776} \\
Uniform & 70.9 & 77.4 & 23.3 & 0.640 \\
NoCap & 69.8 & 75.8 & 22.9 & 0.624 \\
UniformNoCap & 65.7 & 75.1 & 24.9 & 0.601 \\
\bottomrule
\end{tabular}
\vspace{-6mm}
\end{table}

\textbf{Spawn/Goal Sampling:} Similarly, we demonstrate that our spawn/goal sampling, outlined in \Cref{sec:envs}, improves performance. 
We compare training without stair oversampling (Uniform), dropping the geodesic cap in favor of a Euclidean cap (NoCap) and both (UniformNoCap). 
\cref{tab:spawngoal-perf-rom} shows significant and approximately equal drops from removing either, and a larger drop when both components are removed. 
\cref{tab:spawngoal-levels-rom} breaks this down further by splitting the contrast into same-level and cross-level goals. 
The lost performance comes from the cross-level goals. The spawn/goal sampling procedures dramatically help the learning of these complex cross floor navigation behaviors, distinguishing this work from existing legged RL navigation literature. 

\begin{table}[t]
\centering
\caption{Spawn/goal pair success rate by same-level vs. cross-level goals. Positive favors RoM; $\dagger$ marks a 95\% confidence interval excluding zero.}
\label{tab:spawngoal-levels-rom}
\footnotesize
\setlength{\tabcolsep}{4pt}
\begin{tabular}{@{}lcc@{\hspace{1.1em}}cc@{}}
\toprule
& \multicolumn{2}{c}{45\,s} & \multicolumn{2}{c}{120\,s} \\
\cmidrule(lr){2-3}\cmidrule(lr){4-5}
Contrast (\%) & same & cross & same & cross \\
\midrule
RoM $-$ Uniform & $+0.0$ & $+27.8$\rlap{\textsuperscript{$\dagger$}} & $+0.6$ & $+33.5$\rlap{\textsuperscript{$\dagger$}} \\
RoM $-$ NoCap & $-0.6$ & $+30.7$\rlap{\textsuperscript{$\dagger$}} & $-0.2$ & $+37.7$\rlap{\textsuperscript{$\dagger$}} \\
RoM $-$ UniformNoCap & $+1.0$ & $+37.7$\rlap{\textsuperscript{$\dagger$}} & $+0.0$ & $+39.0$\rlap{\textsuperscript{$\dagger$}} \\
\bottomrule
\end{tabular}
\end{table}

\begin{table}[t]
\centering
\caption{Hardware CBF comparison in three environments. Initial conditions and goals are shared across arms and environments.}
\label{tab:e4-hardware}
\footnotesize
\setlength{\tabcolsep}{5pt}
\begin{tabular}{@{}llccc@{}}
\toprule
Obstacle Type & Arm & Success & Collision(s) & Time to goal (s) \\
\midrule
In-Distribution & no CBF & 10/10 & 0/10 & $5.9 \pm 1.5$ \\
 & CBF & 10/10 & 0/10 & $6.7 \pm 2.1$ \\
\addlinespace
Out-of-Distribution & no CBF & 10/10 & 2/10 & $7.5 \pm 1.7$ \\
 & CBF & 10/10 & \textbf{0/10} & $11.0 \pm 5.8$ \\
\addlinespace
Adversarial & no CBF & 10/10 & 4/10 & $5.5 \pm 1.0$ \\
 & CBF & 10/10 & \textbf{0/10} & $6.9 \pm 1.7$ \\
\bottomrule
\end{tabular}
\vspace{-5mm}
\end{table}

\begin{figure*}[t]
\centering
\includegraphics[]{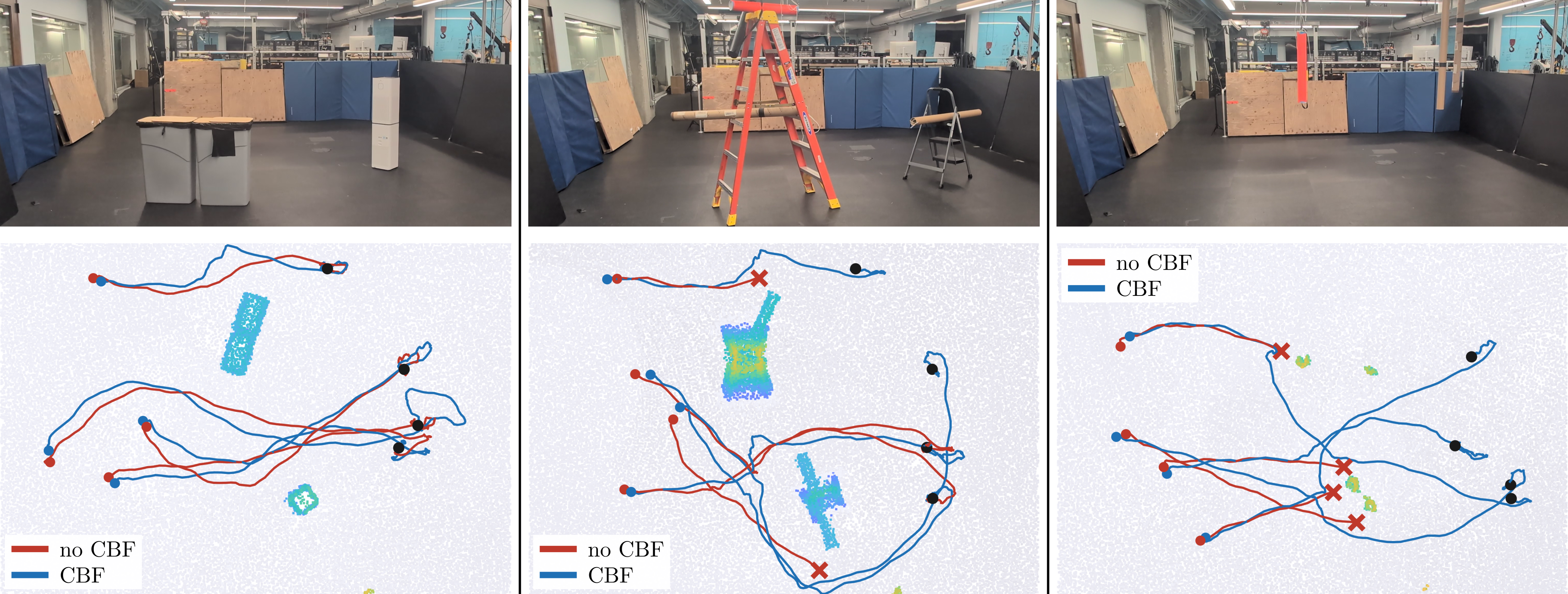}
\caption{Investigation on performance with OOD obstacles. Four of ten trials are plotted (collisions marked with $\times$). (Left) Environment with in-distribution obstacles. (Middle) Environment with OOD obstacles. (Right) Environment with obstacles chosen adversarially, hanging with small LiDAR cross-section.}
\label{fig:cbf}
\vspace{-1mm}
\end{figure*}

\begin{figure*}[t]
\centering
\includegraphics[]{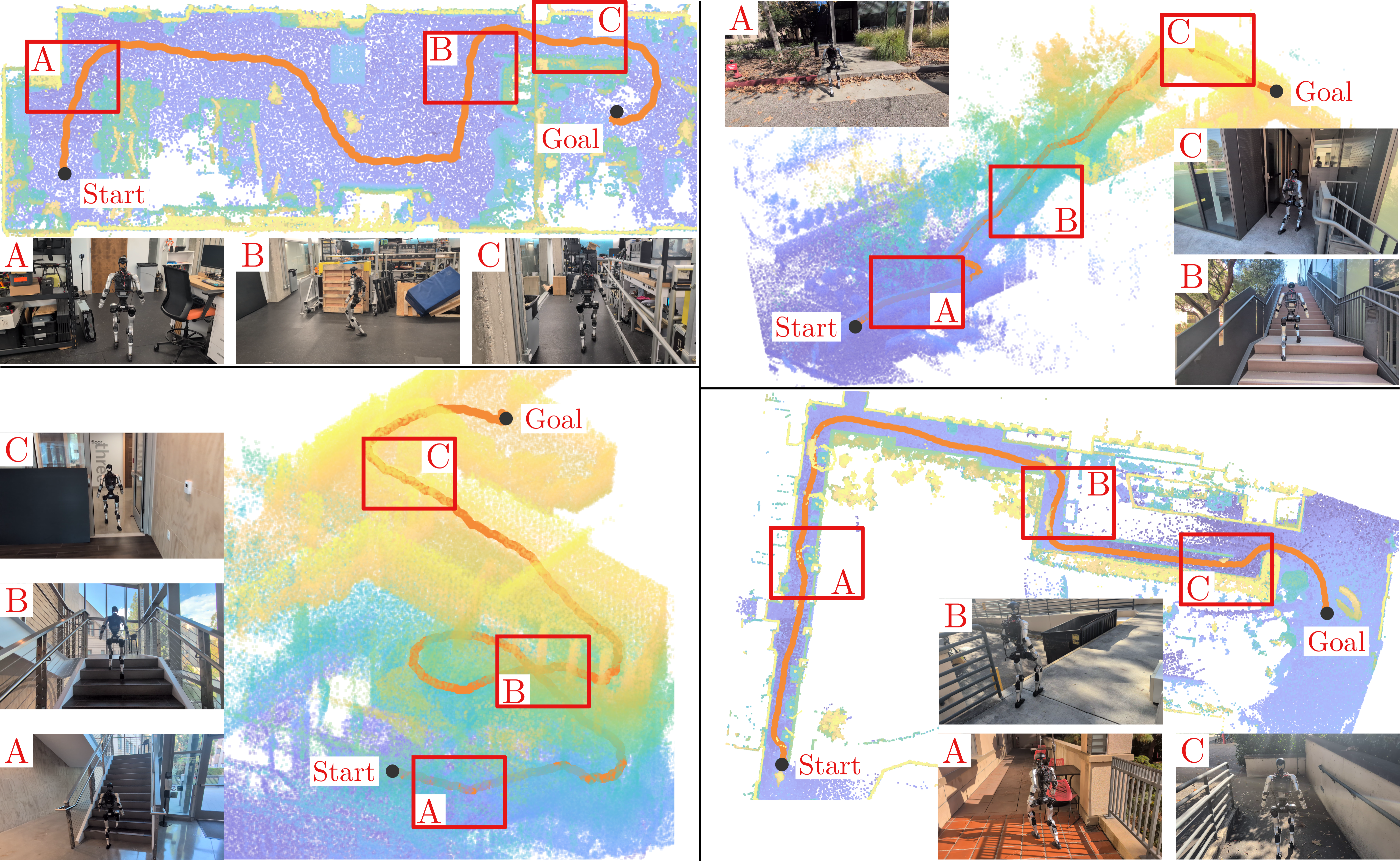}
\caption{Four example deployments in complex, real-world environments. (Top Left) Navigation of a cluttered lab environment, 38\,m path length. (Top Right) 2-story climb (7\,m ascent) on a wire-railed stairwell, 51\,m path length. (Bottom Left) 2-story climb (10\,m ascent) to enter a building from the outside, 51\,m path length. (Bottom Right) A long-horizon outdoor navigation task, avoiding tables, cliff edges and railings, 100\,m path length (1\,m descent).}
\label{fig:hardware}
\vspace{-5mm}
\end{figure*}

\subsection{Safety under Out-of-Distribution Obstacles} \label{sec:cbf_exp}

To demonstrate the sensitivity of the navigation policy to OOD obstacles, we construct challenging environments on hardware.
Training obstacles are geometric, both on the ground and floating, including railings and thin features. 

In-distribution obstacles (\cref{fig:cbf}, left) are chosen to be solid and convex.
The middle pane shows two OOD obstacles, ladders with cardboard tubes sticking out.
Lastly, the right pane contains hanging cardboard tubes, chosen adversarially; thin tubes hung at head height present exceptionally small LiDAR cross-sections. 

Ten spawn/goal points are selected, shared across all environments. Each trial is conducted with the CBF both on and off. To facilitate precise initial condition and goal alignment, the room is mapped using GLIM \cite{koide2024glim}, and the robot is continuously relocalized using a relocalization package built on top of Fast-LIO2 \cite{xu2022fast}. The map is not available to the RoM-Nav policy in any way, which remains strictly mapless. This method is also used in \Cref{sec:hardware}.

The results of this experiment are summarized in \cref{tab:e4-hardware}. 
To avoid cluttering the figure, four of the ten trials (including all collisions) are visualized in \cref{fig:cbf}.
On in-distribution obstacles, we see no collisions in either case.
On OOD obstacles, the CBF makes no collisions, while RoM-Nav makes two on the ladders and four on the hanging tubes.
The CBF pays for its success in collision avoidance by increasing the mean time to goal in the OOD environments. 

\subsection{Hardware Deployment} \label{sec:hardware}

We deploy the RoM-Nav policy on a Unitree G1 robot to perform mapless multi-story long-horizon navigation in real-world environments. 
\cref{fig:hardware} shows four experimental trials on hardware. These trials include vertical climbs of up to 10\,m, which is beyond the training distribution extreme of 8\,m (two 4\,m floors), as well as path lengths of up to 100\,m, which is well beyond the training distribution of 30\,m. None of the hardware trials included a collision. The policy successfully navigates stairwells with thin railings, long outdoor stair climbs, clutter, tables and chairs, and ramps. 

One significant drawback of this policy for real-world deployment is navigating near glass or transparent obstacles. A cardboard sheet is used to block a glass door and window in one trial, which otherwise the robot may have tried to traverse. LiDAR is not a sufficient sensor for navigating transparent obstacles. Other than specific instances of blocking glass, no intentional environmental modifications were required to achieve this navigation performance. Another limitation is the requirement of an accurate goal: placement of an accurate goal in a coherent location, expressed in the body frame, is a strong assumption. Lastly, the planar CBF used in this work could block or close routes the RoM-Nav policy deems traversable, although this is not observed experimentally. Whole-body humanoid safety for navigation and ensuring agreement between the safety filter and navigation policy are potential avenues for future work.

\section{Conclusion}

We investigate autonomous multi-level mapless navigation for humanoid robots. 
Our method first trains a navigation policy on a reduced order model, and uses this policy to kickstart training on a humanoid robot with a frozen locomotion controller.
Nearly all of the gains from the kickstarting process, and also improvements over the single-stage pipeline, occur on trials where the robot must traverse multi-level environments, e.g., to the next story of a building. This is consistent with the literature; the single-stage approach in this paper is heavily based upon prior works that show strong results in single-level navigation. Our extensions show significant benefits for complex multi-story navigation applications on a humanoid robot. 

We also demonstrate the use of a Poisson safety filter to ensure collision avoidance with OOD obstacles on hardware, without reducing navigation success rate. 
LiDAR encoder pretraining and non-uniform spawn/goal sampling procedures were both demonstrated to be critical for cross-level capabilities. 
Finally, the policy is deployed on a Unitree G1 to accomplish safe autonomous real-world multi-floor mapless navigation, covering path lengths over 100\,m and vertical displacements over 10\,m.


\bibliographystyle{IEEEtran}
\balance
\bibliography{main.bib}

\end{document}